\documentclass[lettersize,journal,twoside]{IEEEtran}
\usepackage{amsmath,amsfonts}
\usepackage{algorithmic}
\usepackage{algorithm}
\usepackage{array}
\usepackage[caption=false,font=normalsize,labelfont=sf,textfont=sf]{subfig}
\usepackage{textcomp}
\usepackage{stfloats}
\usepackage{CJKutf8}
\usepackage{url}
\usepackage{multirow}
\usepackage{makecell}
\usepackage{verbatim}
\usepackage{graphicx}
\usepackage{cite}
\usepackage{hyperref}
\usepackage{cleveref}
\usepackage[table,xcdraw]{xcolor}
\usepackage{booktabs}
\usepackage[flushleft]{threeparttable}
\usepackage{subfloat}
\begin{document}

\title{Voxel-based Point Cloud Geometry Compression with Space-to-Channel Context}

\author{Bojun Liu, Yangzhi Ma, Ao Luo, Li Li,~\IEEEmembership{Member,~IEEE,}  Dong Liu,~\IEEEmembership{Senior Member,~IEEE}  

        % <-this % stops a space
\thanks{Date of current version December 6, 2024}% <-this % stops a space

\thanks{B. Liu, Y. Ma, L. Li, and D. Liu are with
University of Science and Technology of China, Hefei 230027, China (e-mail: liubj@mail.ustc.edu.cn; 
mayz@mail.ustc.edu.cn;
lil1@ustc.edu.cn; 
dongeliu@ustc.edu.cn).}

\thanks{A. Luo is with the School of Fundamental Science and Engineering, Waseda University, Shinjuku-ku, Tokyo, 169-8555, Japan (email: luo.ao@toki.waseda.jp).
}
% \thanks{L. Zhang is with the Advanced Video Group (AVG), Bytedance Inc., San Diego, CA 92122 USA (email: lizhang.idm@bytedance.com).}
}

% \markboth{Submitted to IEEE transactions on image processing}
% {Liu \MakeLowercase{\textit{et al.}}: Voxel-based Point Cloud Geometry Compression with Space-to-Channel Context} 

\maketitle
\begin{abstract}
Voxel-based methods are among the most efficient for point cloud geometry compression, particularly with dense point clouds. However, they face limitations due to a restricted receptive field, especially when handling high-bit depth point clouds. To overcome this issue, we introduce a stage-wise Space-to-Channel (S2C) context model for both dense point clouds and low-level sparse point clouds. This model utilizes a channel-wise autoregressive strategy to effectively integrate neighborhood information at a coarse resolution. For high-level sparse point clouds, we further propose a level-wise S2C context model that addresses resolution limitations by incorporating Geometry Residual Coding (GRC) for consistent-resolution cross-level prediction. Additionally, we use the spherical coordinate system for its compact representation and enhance our GRC approach with a Residual Probability Approximation (RPA) module, which features a large kernel size. Experimental results show that our S2C context model not only achieves bit savings while maintaining or improving reconstruction quality but also reduces computational complexity compared to state-of-the-art voxel-based compression methods.

\end{abstract}


\section{Introduction}
\IEEEPARstart{P}{oint} cloud is a type of 3D spatial representation consisting of a large number of discrete points, which has been widely applied across various fields like computer graphics, autonomous driving, virtual reality, and augmented reality. Point cloud data offers high resolution and precision, which is beneficial for 3D modeling and spatial analysis. However, these qualities also result in substantial data volumes, which pose significant challenges for efficient storage and transmission, thereby necessitating the development of efficient point cloud geometry compression methods~\cite{krivokuca2019volumetric, tzamarias2022fast, thanou2016graph,de2016compression} to support practical applications. \par
Recently, the Moving Pictures Experts Group (MPEG) has introduced the Geometry-based Point Cloud Compression (G-PCC) \cite{g-pccmpeg} standard for point cloud compression. G-PCC employs the octree \cite{jackins1980oct, meagher1982geometric, schnabel2006octree} structure to organize point clouds, with the total number of levels in the octree determined by the bit depth. This hierarchical structure enables G-PCC to apply the hand-crafted rules for extracting neighboring context within two adjacent levels. Consequently, the point cloud can be compressed and reconstructed in a level-wise autoregressive manner. Drawing inspiration from the efficient representation of the octree, the latest deep learning-based point cloud compression methods \cite{luo2024scp, wang2022sparse} aim to develop the learned entropy model and have achieved the state-of-the-art compression performance by capturing complex patterns and dependencies within the point cloud data.  \par
Octree-based learned entropy models \cite{fu2022octattention,song2023efficient}, explicitly construct an octree and optimize the network parameters to estimate symbol distributions of octree node. These methods leverage the attention mechanism \cite{vaswani2017attention} to capture long-distance dependencies, thereby exploring the correlation between nodes. This approach is particularly suitable for processing large-scale sparse LiDAR point clouds. However, for dense object point clouds, which consist of millions of points with high density, long-distance dependencies are not essential. In such cases, the heavy computation associated with these models becomes a liability. 
\par
Instead, dense object point cloud can be effectively processed by voxel-based learned entropy model~\cite{wang2021multiscale,wang2022sparse}, which implicitly utilizes the cross-level occupancy prediction structure of octree. This method employs 3D sparse convolution~\cite{choy20194d} to analyze the voxelized point cloud and incorporates an eight-stage autoregressive occupancy prediction to capture neighborhood information at each level efficiently. However, as the bit depth increases, the receptive field of sparse convolution decreases, which limits the effectiveness of the voxel-based point cloud compression method on high-bit depth data that has high position precision.\par
%可学习的八叉树处理稀疏点云, 稀疏卷积的voxel方法处理稠密点云
%voxel方法的优缺点, 我们针对性的改进.
We analyze the resolution constraints of voxel-based point cloud compression methods from two factors. Firstly, the cross-level occupancy probability prediction module uses an upsampling operation to align the resolutions of adjacent levels. However, this upsampling process leads to a reduction in the receptive field of sparse convolution during entropy model inference, thereby limiting the integration of local geometry information for accurate cross-level predictions. Secondly, the eightfold upsampling is also utilized to increase the number of points. While this approach is effective for dense point clouds, it is unsuitable for sparse point clouds, particularly in the last few octree levels where sparsity is extreme and the number of points does not increase. In this case, upsampling becomes a redundant operation, increasing the computational complexity.
% We analyze the bit rate consumption at each level of the voxel-based method \cite{wang2022sparse} on sparse LiDAR point cloud dataset Ford \cite{pandey2011ford} and depict the results in \Cref{introduction}. It is observed that higher levels account for the majority of the bit rate, with the bit rate for these levels approaching approximately 3 bits per point (bpp), corresponding to the coordinates~($\Delta x, \Delta y, \Delta z$) of a point at that level. This trend can be attributed to two factors. Firstly, the number of points at the higher levels is nearly equivalent to that in the original point clouds. Consequently, the bit rate at these levels is allocated to encode the residual positions~($\Delta x, \Delta y, \Delta z$) to refine the coordinates' positions.
% Secondly, the receptive field of sparse convolution with a fixed kernel size decreases as the level increases, which limits its ability to further reduce the bit rate from the original 3 bpp by exploiting the correlations within the neighborhood. 
\par 
In this paper, our basic idea is to enhance the analytical capabilities of sparse convolution for the voxel-based entropy models to extract large-scale context. Subsequently, we propose a stage-wise Space-to-Channel (S2C) context model for dense point clouds and low-level sparse LiDAR point clouds. This model shifts the point-wise autoregressive method in \cite{wang2022sparse} to a channel-wise autoregressive one by storing the stage-wise occupancy information in the channels, thereby expanding the receptive field to fuse neighborhood information with the same kernel size. For high-level sparse point clouds, we further propose a level-wise Space-to-Channel (S2C) context model that addresses resolution limitations by incorporating the Geometry Residual Coding (GRC) module. This module converts the increase in spatial resolution to an expansion in the channel dimension of the sparse tensor by storing the level-wise residual positions within the tensor's channels. Moreover, we opt to transform the Cartesian coordinate system into a Spherical coordinate system, which offers a more compact positional representation. This transformation enables the application of our proposed GRC at an earlier level. Additionally, we propose a Residual Probability Approximation (RPA) module with an enlarged kernel size to efficiently extract a large-scale context. 
\par

Our contributions as presented in this paper can be summarized as follows:
\begin{itemize}
    \item We propose the Space-to-Channel (S2C) context extraction method for voxel-based entropy model in point cloud geometry compression, enhancing the ability of sparse convolution to efficiently capture and integrate broader neighborhood correlations.
    \item We develop two distinct context models (stage-wise and level-wise) for dense and sparse point clouds, simplifying the autoregressive approach by extending it into the channel dimension and enabling the extraction of large-scale context more effectively.
    \item We conduct a comprehensive set of experiments on both dense and sparse point clouds to validate the performance of our proposed S2C context model in the point cloud geometry compression. Experimental results demonstrate the effectiveness of our method in terms of bit savings and reduced encoding/decoding complexities.
\end{itemize}

The remainder of this paper is organized as follows. In \Cref{sec:related_work}, we review related work on octree-based and voxel-based point cloud compression. \Cref{sec:method} presents the details of our proposed stage-wise and level-wise S2C context model. In \Cref{sec: experiments}, we describe the configuration of the experiments and analyze the results.  \Cref{sec: conclusion} concludes this paper.

%改进与描述
%贡献

\section{Related Work}
\label{sec:related_work}
\subsection{Octree-based point cloud compression}
%octsqueeze, voxelnet, octsttention,ehem,scp
\label{octree}
Octree \cite{jackins1980oct} is a highly efficient data structure for organizing point clouds, which recursively divides the 3D space into eight uniform sub-spaces at each bit level and arranges them into a sequence in pre-defined order, such as Morton order and Hilbert order. Traditional point cloud codecs~\cite{thanou2016graph,schwarz2018emerging,garcia2019geometry,cao2021compression} utilized octree to represent the point cloud and designed the context model based on handcrafted rules. In recent years, learned entropy models for point cloud compression have emerged and evolved, demonstrating superior rate-distortion performance over traditional methods. \par
Huang \textit{et al.}~\cite{huang2020octsqueeze} designed the first learned entropy model based on the octree structure by modeling and encoding the octree symbols, which utilized a multi-layer perceptron to integrate the contextual information from the ancestor nodes. 
Based on this pioneering work, Que \textit{et al.}~\cite{que2021voxelcontext} leveraged local voxel context to compress the static and dynamic octree structured data. They also developed a coordinate refinement module for post-processing the point cloud, which resulted in high-quality reconstructed coordinates of the leaf nodes.
Fu \textit{et al.}~\cite{fu2022octattention} further gathered the information from sibling and ancestor nodes to remove the spatial redundancy. They employed a large field context model to improve the compression performance. However, this approach resulted in an unacceptable serial decoding time. 
To further explore the effectiveness of large-scale context and reduce the decoding time, Song \textit{et al.}~\cite{song2023efficient} introduced a hierarchical attention structure that expanded the receptive field from 1024 to 8192 modes. Additionally, they proposed a grouped context structure, enabling efficient group-based parallel decoding. 
Luo \textit{et al.}~\cite{luo2024scp} innovatively transformed point clouds from Cartesian to Spherical coordinates, fully leveraging the physical prior inherent in the captured mode of LiDAR point clouds.\par
Octree-based methods have yielded state-of-the-art rate-distortion performance for compressing sparse point clouds. However, their high computational complexity and decoding latency present significant limitations for practical implementation.

\subsection{Voxel-based point cloud compression}
\label{voxel}
%fredic %departion %pcgcv2 %grasp-net %sparse-pcgc
In some pioneering work~\cite{quach2019learning,quach2020improved,wang2021lossy, gu20193d}, voxel-based methods voxelized point clouds and applied 3D convolutions to process them into a compact latent representation. They reconstructed the point clouds by predicting the occupancy probability of voxels. Nevertheless, these approaches incur significant inefficiencies due to the extensive processing of numerous empty voxels. Wang \textit{et al.}~\cite{wang2021multiscale} applied 3D sparse convolutions~\cite{choy20194d} for the progressive estimation of voxel occupancy. By leveraging a multiscale structure to represent the voxelized point cloud, they enabled efficient feature analysis and aggregation on the occupied voxels. The proposed sparse convolution-based framework, operating solely on non-empty voxels, demonstrated memory efficiency and enhanced performance in dense point clouds. 
Pang \textit{et al.}~\cite{pang2022grasp} proposed a heterogeneous approach for lossy point cloud compression, which included a coarse point cloud representation and an enhancement layer. The approach combined traditional codec for the coding of coarse representation, with an enhancement layer that utilizes a point-based network and sparse convolution module to refine the geometric details. 
Wang \textit{et al.}~\cite{wang2022sparse} further presented a multi-scale sparse tensor representation, significantly advancing the field of sparse convolutions for point cloud compression. Their method described the binary status (occupancy or empty) of the voxel and employed a Sparse convolution-based Occupancy Probability Approximation (SOPA) model to estimate the occupancy probabilities. This approach has demonstrated its capability to achieve state-of-the-art performance across both lossless and lossy compression modes, particularly for dense point clouds.
\par 
The sparse convolution-based method is a low-complexity solution for point cloud processing which is attractive to practical applications.
However, the approach might not be optimal for point clouds with higher voxel resolutions at greater bit depths, due to the limitations of its receptive field.

\section{The Proposed Method}
\label{sec:method}
In this section, we begin by analyzing the resolution constraint inherent in voxel-based methods. 
Following this, we present our proposed stage-wise Space-to-Channel (S2C) context model. Compared to the 8-stage point-wise autoregressive context model, our channel-wise autoregressive context model captures a wider receptive field and deeply mines the correlation information between neighboring voxels. Moreover, we propose the level-wise Space-to-channel context model for high-level coding of sparse point clouds. This model expands the feature channels of sparse tensors to prevent resolution from increasing as levels deepen. It consists of three main components: geometry residual coding, coordinate system selection, and a residual probability approximation module.                                                                                                                                            

\subsection{Voxel-based Coding Method Analysis}
\label{sec:analysis}

The latest voxel-based method~\cite{wang2022sparse} 
introduces a multi-level voxelization approach that mirrors an octree structure, achieved through bit-wise slicing of the point cloud. For instance, a point cloud with an 18-bit depth is decomposed into 18 levels, with resolution progressively increasing at higher levels. This process effectively represents the point cloud in a hierarchical series of voxels, with each level encapsulating the geometric information of its predecessor. The method employs sparse tensors to organize the voxels at each level efficiently and incorporates a novel SparseCNN-based Occupancy Probability Approximation (SOPA) model for estimating the occupancy probability of the next level. In~\cite{wang2022sparse}, SOPA upsamples the voxels at the current level into eight sub-voxels, aligning them with the resolution of the next level and thereby enabling the prediction of occupancy probabilities for the latter. The actual occupancy information of the next level is then encoded and decoded based on these predictions. Following decoding, only occupied voxels are retained, with empty ones being pruned away. The entire process at each level is divided into eight stages, employing an autoregressive approach to progressively predict the eight sub-voxels derived from a single voxel.
\par

This upsample-then-prune methodology, while inherently aligned with the increasing resolution characteristic of an octree structure, faces constraints when applied via sparse convolution. The upsampling process results in a base unit voxel of diminished volume. Consequently, sparse convolution with a fixed kernel size experiences a reduced receptive field at higher resolutions as opposed to coarser resolutions, thereby constraining its ability to extract correlation information among neighboring voxels for dense point clouds and low-level sparse point clouds.\par

\begin{figure}[t]
\centering
\centerline{\includegraphics[width=0.90\linewidth]{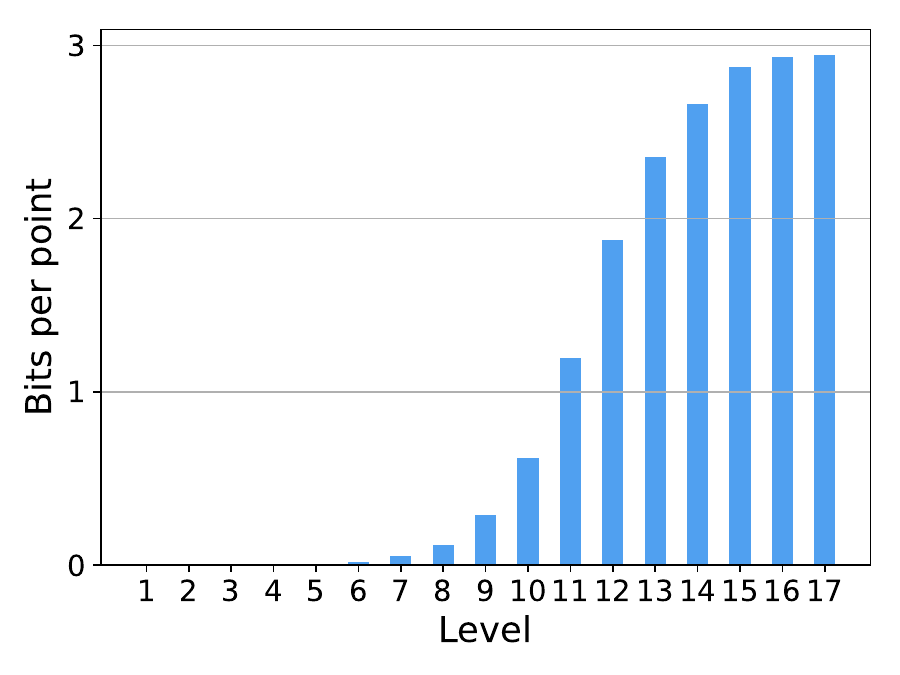}}
\caption{The bit rate consumption across each level of the voxel-based method~\cite{wang2022sparse} on the Ford dataset. The bit rate for the initial few levels is approximately 0, while the bit rate for the highest levels tends to stabilize at 3 bits per point.}
\label{introduction}
\end{figure}

This problem becomes particularly pronounced when it comes to processing high-level sparse LiDAR point clouds. We analyze the bit rate consumption at each level of the voxel-based method \cite{wang2022sparse} on sparse LiDAR point cloud dataset Ford \cite{pandey2011ford} and depict the results in Fig.~\ref{introduction}. It is observed that higher levels account for the majority of the bit rate, with the bit rate for these levels approaching approximately 3 bits per point (bpp), corresponding to the coordinates~($\Delta x, \Delta y, \Delta z$) of a point at that level. This trend can be attributed to two factors. Firstly, the number of points at the higher levels is nearly equivalent to that in the original point clouds. Consequently, the bit rate at these levels is allocated to encode the residual positions~($\Delta x, \Delta y, \Delta z$) to refine the coordinates' positions.
Secondly, the receptive field of sparse convolution with a fixed kernel size decreases as the level increases, which limits its ability to further reduce the bit rate from the original 3 bpp by exploiting the correlations within the neighborhood. \par 

% Firstly, sparse LiDAR point clouds, characterized by a sparser distribution and higher bit depth, initially experience an increase in point count upon upsampling at the initial levels. However, as the level increases, the number of points can nearly match the original point cloud, suggesting that the upsampling operation may become redundant and inefficient at these higher levels. 
% Secondly, as the level increases, the coordinate resolution expands, leading to a relative decrease in the effective receptive field of the sparse convolution. This reduction in effective field imposes constraints on the sparse CNN's ability to capture the spatial correlations between central points and their neighbors at high levels.
\begin{figure*}[thb]
\centering
\centerline{\includegraphics[width=0.95\linewidth]{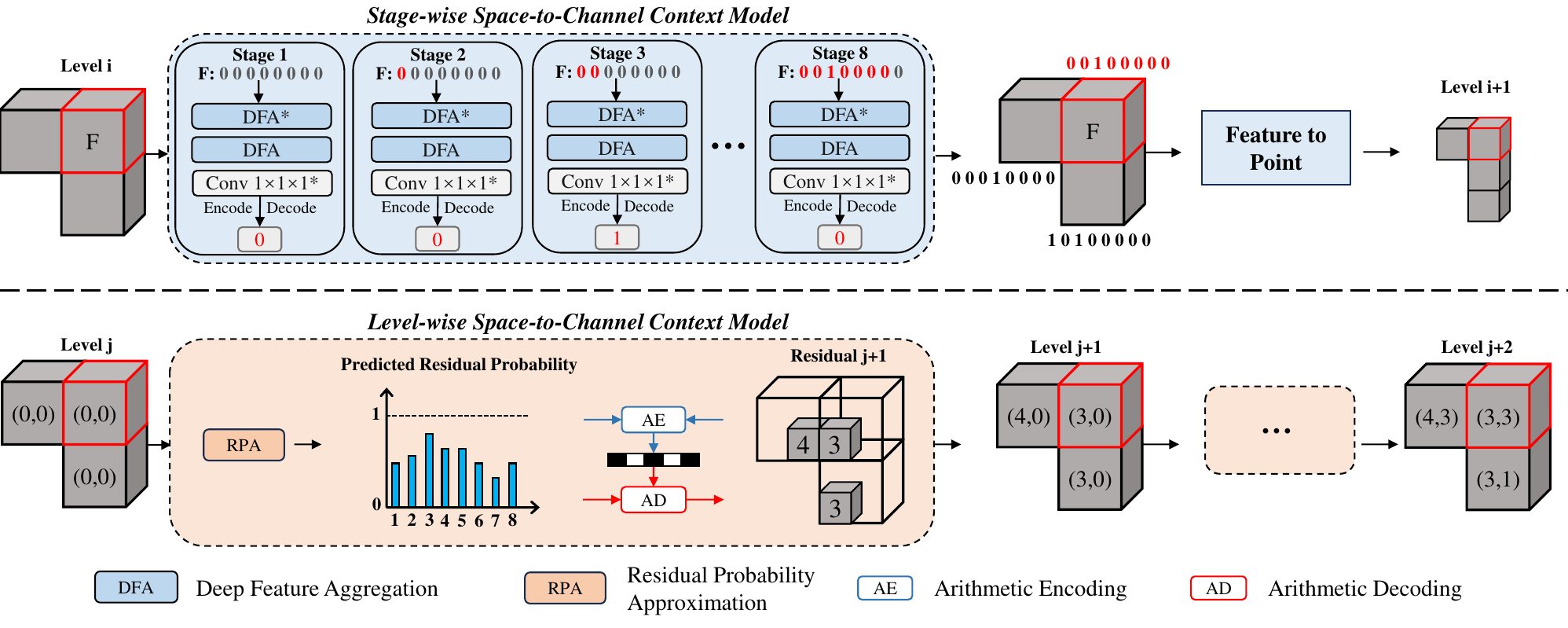}}
\caption{\textbf{Top:}~The architecture of stage-wise Space-to-Channel (S2C) context model from level $i$ to level $i+1$. The model bypasses upsampling by converting the sub-space of level $i+1$ to the channel of level $i$. At each stage, the occupancy probability for each sub-voxel is predicted and encoded, allowing the voxels at level $i+1$ to be reconstructed from the decoded feature channels of level $i$ through the feature-to-point module. The symbol $*$ denotes that the corresponding module employs parameter sharing across all stages.
\textbf{Bottom:}~The architecture of level-wise Space-to-Channel (S2C) context model from level $j$ to level $j+2$. The residual position probability of level $j+1$ relative to level $j$ is predicted using a Residual Probability Approximation (RPA) model. Based on the predicted probability distribution, the relative residual position is encoded and decoded. The same iterative process is then applied from Level $j+1$ to $j+2$.
The red boxes denote the resolution of voxels at each level. The receptive field of sparse convolution remains constant throughout the inference process in our proposed S2C context model.
}
\label{framework}
\end{figure*}

% \begin{figure*}[t]
% \centering
% \centerline{\includegraphics[width=0.95\linewidth]{figs/crop_stage-wise.pdf}}
% \caption{Stage-wise Space-to-Channel (S2C) context model for dense point clouds. The model bypasses up-sampling by converting the sub-space of level $i+1$ to the channel of level $i$. At each stage, the occupancy probability for each sub-voxel is predicted and encoded, allowing the voxels at level $i+1$ to be reconstructed from the decoded feature channels of level $i$.  
% }
% \label{stage-wise}
% \end{figure*}

\subsection{Stage-wise Space-to-Channel Context Model}
\label{dense_network}
To preserve the receptive field of sparse convolution within the context model, we propose to transform the increase of space resolution into an expansion of the feature channel dimensions within sparse tensor, thereby eliminating the need for upsampling operations. Following the model inference for occupancy probability prediction, we perform the feature-to-point transformation to reconstruct the point clouds at the next level. \par
In this work, we propose a novel stage-wise (S2C) context model that employs a channel-wise autoregressive approach to predict the occupancy probability of sub-voxels. As shown at the top of Fig.~\ref{framework}, the feature at level $i$ is expanded into eight channels, each corresponding to one of the eight sub-voxels at level $i+1$.  At stage 1, each channel is initialized to 0. We then predict the occupancy probability of the first sub-voxel. Based on this prediction, the occupancy state is encoded and decoded. With the true occupancy state of the first sub-voxel, we update the initialization of the first channel. This updated feature is then used to predict the occupancy of the next sub-voxel. This process repeats for eight stages, allowing us to encode and decode from level $i$ to level $i+1$. Finally, we are able to reconstruct the voxels at level $i+1$ by leveraging the feature channels from level $i$ through a feature-to-point transformation. This process establishes a one-to-one mapping between sub-voxels and channels.
\par
The sparse CNN-based prediction module is composed of two Deep Feature Aggregation (DFA) modules introduced in~\cite{wang2022sparse} and a series of sparse convolutions with a kernel size of 1. The DFA module consists of three Inception-ResNet modules (IRN, detailed in Fig.~\ref{backbone}) and two layers of sparse convolution. To ensure our method has a similar number of parameters to the approach described in~\cite{wang2022sparse}, we share the parameters of the first DFA module and the convolution layer at each stage. By transforming the spatial information of the sub-voxels into channel information for voxels, we effectively integrate neighborhood information. Notably, our approach expands only the channel dimension of the input tensor, thereby maintaining model inference efficiency. Furthermore, operations on sparse tensors at level $i$ encompass a broader receptive field compared to those at a higher level. Moreover, the number of processed voxels remains constant across stages, contributing to a reduction in the computational complexity.

\subsection{Level-wise Space-to-Channel Context Model}
Our proposed stage-wise S2C context model is designed to enhance the compression of dense point clouds and low-bit levels of sparse point clouds, which contain a large number of points confined within a limited spatial range. However, this approach faces challenges when extended to high-bit levels of sparse LiDAR point clouds, as the receptive field even at level $i$ is insufficient to capture the necessary local geometry information. To address this, our basic idea is to maintain the resolution of sparse tensors at a lower level during the processing of high-bit levels of sparse point clouds. Consequently, we further propose a level-wise Space-to-Channel context model that encompasses three key components: geometry residual coding, coordinate system selection, and residual probability approximation module. 

\subsubsection{\textbf{Geometry Residual Coding}}
\label{sec:Residual coding}
To alleviate the resolution constraint and simplify the cross-level prediction, we propose the Geometry Residual Coding (GRC) method, which leverages the inherent physical prior of sparse LiDAR point cloud distributions. As shown at the bottom of Fig.~\ref{framework}, with the assumption that the number of points at a high level $j$ has matched the original point cloud, the voxel at level $j+1$ is identified as the sole sub-voxel of its counterpart at level $j$. Consequently, the prediction task is fundamentally changed. In~\cite{wang2022sparse}, the task involves predicting the occupancy probability for eight sub-voxels of each voxel after upsampling, which has 254 states (excluding the empty state). In contrast, our GRC simplifies this by predicting the position of the sole sub-voxel without upsampling, thus reducing the states to just 8. The compression of the state space allows for more precise probability prediction.
\par
Based on the above, we develop a Residual Probability Approximation (RPA) model to predict the distribution of the residuals between consecutive levels, that is, from level $j$ to level $j+1$. Using the predicted probability distribution, the residual is encoded and then decoded. Removing the upsampling step leads to faster inference. After the bitstream is decoded, the residual is stored in a feature channel of the sparse tensor without changing the coordinate. This ensures that the coordinates at level $j+1$ are the same as those at level $j$, thus preventing an increase in resolution. The residual of level $j+2$ is stored in the next feature channel in the same way. This design is robust for the increases in coordinate resolution. As illustrated in Fig.~\ref{framework}, the GRC can effectively capture a large-scale context about the neighboring voxels with a larger receptive field.\par

% \begin{figure*}[t]
% \centering
% \subfloat[KITTI-Cartesian]{\includegraphics[width=1.3in]{figs/kitti_cer.pdf}%
% \label{kitti_cer}}
% \hfil
% \subfloat[KITTI-Spherical]{\includegraphics[width=1.3in]{figs/kitti_sph.pdf}%
% \label{kitti_sph}}
% \hfil
% \subfloat[Ford-Cartesian]{\includegraphics[width=1.3in]{figs/ford_cer.pdf}%
% \label{ford_cer}}
% \hfil
% \subfloat[Ford-Spherical]{\includegraphics[width=1.3in]{figs/ford_sph.pdf}%
% \label{ford_sph}}
% \hfil
% \caption{The Bird’s-eye-view visualization of point clouds No. 000000 in Sequence 11 (a,~b) of KITTI~\cite{behley2019semantickitti} dataset and No. 0100 in Sequence 2 (c,~d) of Ford~\cite{pandey2011ford} dataset. We show the distribution of point cloud in $x-y$ plane for Cartesian space (a,~c) and $\rho-\phi$ plane for Spherical space (b,~d).}
% \label{fig:visual}
% \end{figure*}

We optimize the proposed GRC by the cross-entropy between the predicted residual probability and the actual residual on each voxel, which is defined as:
\begin{align}
\mathcal{L} &= -\sum\limits_{i}\log{\tilde{p}_i(x_i)},
\end{align} 
where $\tilde{p}_i(x_i)$ is the distribution of residual $x_i$ predicted by the proposed RPA model.
The final coordinates of the point cloud can be reconstructed by converting the features within the sparse tensor into real residual coordinates. Specifically, the decoded residual from the $n$-th channel, denoted as $R_n$ and ranging from 1 to 8, is initially binarized into 3 bits, $r_{n,1},r_{n,2},r_{n,3}$, corresponding to the three dimensions of the coordinate system. Subsequently, the reconstructed coordinate is calculated by taking the base coordinate from level $j$, adding the products of the residuals from each channel and their corresponding weights. This process can be mathematically formulated as:
\begin{gather}
r_{n,1},~r_{n,2},~r_{n,3} = Binary(R_n-1),
\\
C_{j+m,d} = C_{j,d} + \sum\limits_{n=1}^{m}2^{m-n}r_{n,d},~d \in \{1,2,3\},
\end{gather}
where $C_{j,d}$ represents the coordinate of $d$-th dimension at the level $j$. $m$ denotes the total number of levels that apply the GRC method. \par
We integrate our proposed GRC into the overall compression scheme. As depicted in Fig.~\ref{framework}, the initial phase of the process uses our proposed stage-wise S2C context model to quickly increase the number of points at low levels. Once the point count at level $j$ is close to that of the original point cloud, GRC is applied to refine the coordinates via residual encoding.
In practice, the number of points at level $j$ might be slightly fewer than that in the original point cloud. The extra points from the original are written directly into the bitstream. Overall, the transmitted bitstream of the proposed scheme consists of three parts: low-level coarse representation, geometry residual, and directly encoded points.

\subsubsection{\textbf{Coordinate System Selection}}
\label{sec:spher}
\begin{figure}[tb]
\centering
\centerline{\includegraphics[width=0.95\linewidth]{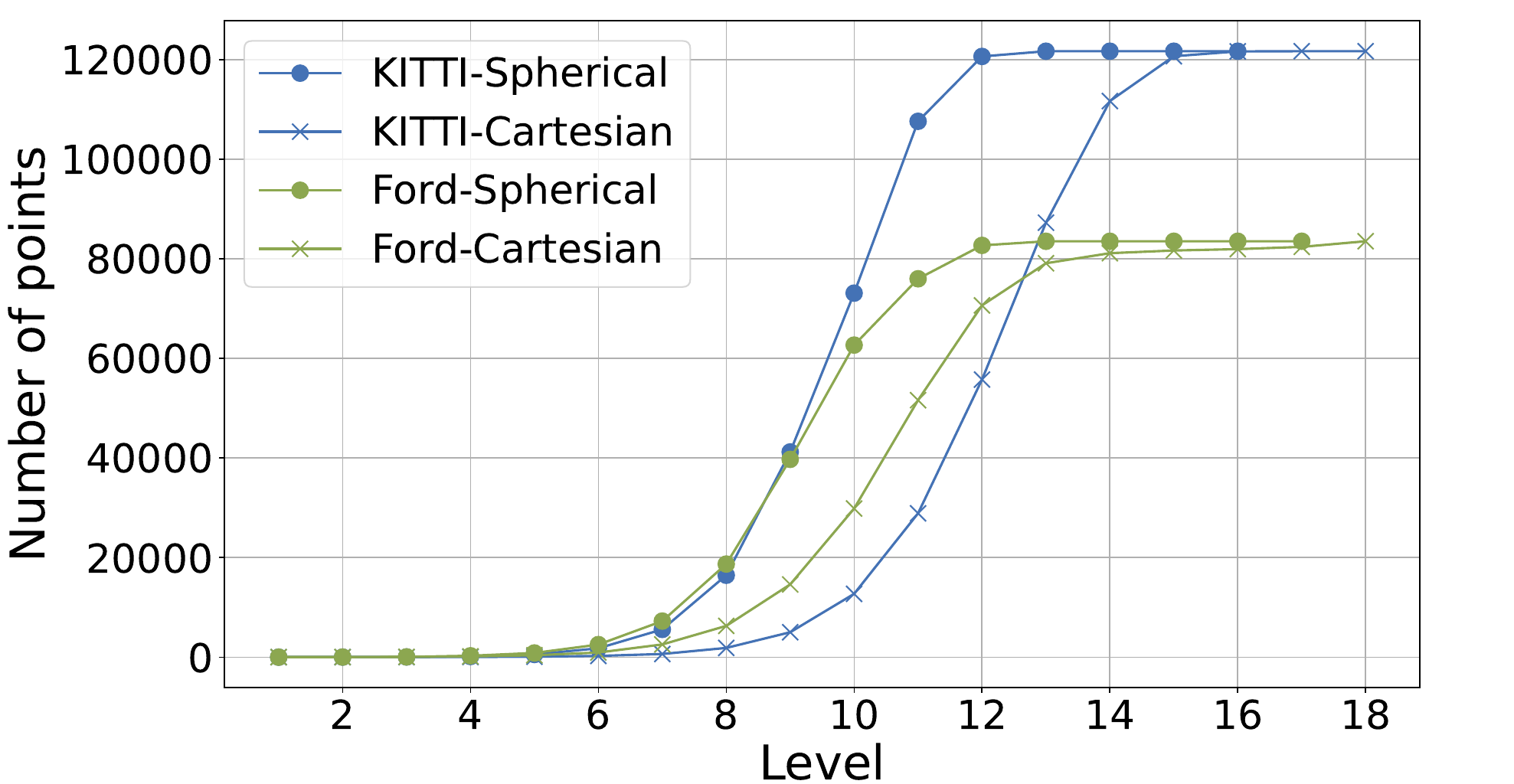}}
\caption{The number of points across levels in Cartesian and Spherical coordinate systems for the KITTI~\cite{behley2019semantickitti} and Ford~\cite{pandey2011ford} datasets. The Cartesian coordinate space exhibits a maximum of 18 levels for both KITTI and Ford datasets. In contrast, the Spherical coordinate space reaches a maximum of 16 levels for KITTI 
and 17 for Ford.
}
\label{distribution}
\end{figure}

Due to the rotational acquisition nature of LiDAR point clouds, several studies~\cite{luo2024scp,thomas2018semantic,lei2020spherical,zhang2020polarnet} have attempted to utilize this physical prior for processing sparse LiDAR point clouds. 
% Luo \textit{et al.}~\cite{luo2024scp} introduced a novel approach by initially converting the Cartesian coordinate $(x,y,z)$ of point cloud into Spherical coordinate $(\rho,\theta,\phi)$, thereby enhancing the octree-based entropy model's capacity to extract informative context. 
The transformation from Cartesian $(x,y,z)$ to the Spherical coordinate system $(\rho,\theta,\phi)$ is mathematically expressed as follows: 
\begin{equation}
\left\{
\begin{aligned}
    \rho &= \sqrt{x^2+y^2+z^2}, \\
    \theta &= \arccos{\frac{z}{\sqrt{x^2+y^2+z^2}}}, \\
    \phi &= \arctan\frac{y}{x},
\end{aligned}
\right.
\end{equation}
where $\rho$ represents the radial distance, $\theta$ is the elevation angle, and $\phi$ is the azimuth angle.  
% As illustrated in~\Cref{fig:visual}, we present the distribution of the LiDAR point cloud across Cartesian and Spherical coordinate spaces. The visualization reveals that the point cloud in Spherical space exhibits a denser distribution with a more limited range in comparison to the Cartesian space. Furthermore, the circular lines in Cartesian space are projected into straight lines within the Spherical coordinate system. This transformation suggests that the Spherical coordinate system provides a more compact and energy-efficient representation of the sparse LiDAR point cloud. \par
% Additionally, we quantize the point cloud datasets following the setting of~\cite{lei2020spherical} and analyze the distribution of points across various levels. A detailed explanation of the quantization process can be referenced in \cite{luo2024scp}. 
We statistic the number of points in sparse point clouds at each level in Cartesian and Spherical coordinate systems across two datasets and present the results in Fig.~\ref{distribution}. It is observed that there is a rapid increase in the number of points starting from level 6. By level 15 in the Cartesian coordinate system, the number of points matches the count of the original point clouds. In contrast, the Spherical coordinate system achieves a comparable scale in point count by level 12. This discrepancy arises because uniform partitions in Spherical space $(\rho, \theta, \phi)$ lead to non-uniform partitions in Cartesian space $(x,y,z)$, which tend to be denser near the origin and sparser at greater distances. Consequently, the Spherical coordinate system facilitates a more rapid increase in point count. This characteristic aligns well with our proposed GRC method, which relies on the assumption that the number of points at a high level $j$ corresponds to that of the original point cloud. By fixing the resolution at level 12, we enable the execution of GRC across multiple high levels within Spherical space. We choose the Spherical coordinate system for representing the point cloud due to its effectiveness in capturing a large-scale local geometry context for GRC.

\begin{figure*}[t]
\centering
\centerline{\includegraphics[width=0.95\linewidth]{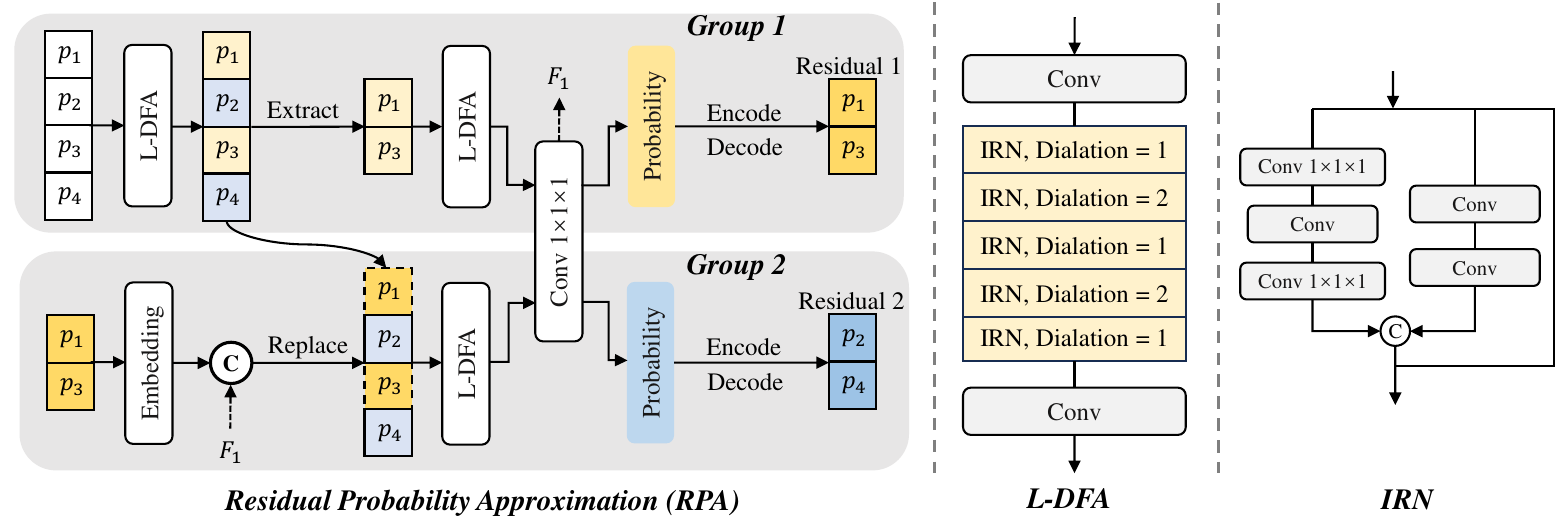}}
\caption{\textbf{Left:}~The illustration of the proposed Residual Probability Approximation (RPA) model. The model begins by analyzing the input sparse tensor with the Large-scale deep feature aggregation module. It then divides the tensor into two groups (\textcolor{yellow}{Group 1} and \textcolor{blue}{Group 2}). The first group's features are used to estimate the probability distribution of its voxel residuals. The second group's probability distribution is predicted using both its own features and the encoded information from the first group's voxels.
\textbf{Middle:}~Overview of the Large-scale Deep Feature Aggregation (L-DFA) architecture. This module employs a large kernel and dilated convolutions to broaden the receptive field, building upon the DFA module used in~\cite{wang2022sparse}.
\textbf{Right:}~The architecture of Inception-ResNet (IRN), which is the basic module of our proposed L-DFA.
}
\label{backbone}
\end{figure*}
% Projecting voxels of identical volume from Spherical to Cartesian space reveals that voxels with smaller radial distances are smaller in size compared to those with large radial distances. 

% This partitioning strategy aligns more closely with the natural distribution of LiDAR point clouds, 

\subsubsection{\textbf{Residual Probability Approximation}}
\label{network}
%点数分布，及球坐标可视化，选择坐标系，球坐标变换及量化公式。
%大核backbone，空洞卷积   
We introduce the Residual Probability Approximation (RPA), a grouped residual context model designed to efficiently aggregate neighbor voxel information. As illustrated in Fig.~\ref{backbone}, the process initiates with the input sparse tensor being analyzed by our proposed Large-scale Deep Feature Aggregation (L-DFA) module, which gathers information from the local region. The sparse tensor is then categorized into two sequences based on their indices: odd and even groups. The voxels of the first group are further analyzed in depth by another L-DFA module to extract deeper features. Following this, the features from the first group are used to estimate the probability distribution of the residuals for these voxels through a series of sparse convolutions with a kernel size of 1. Concurrently, the convolutional layer also generates an intermediate feature map $F_1$, which aids in the prediction process for the second group of voxels. 
\par 
Following the encoding or decoding of the first group's residuals, the information of group 1 can be utilized as the neighboring prior. Initially, the residuals of group 1 are transformed through one-hot encoding into an ensemble of vectors with eight channels. These vectors are then embedded into latent representations by a sparse convolution layer with a large kernel size. Subsequently, the latent features are concatenated with the intermediate feature $F_1$ of group 1, effectively updating the feature of the first group's voxels in the initially aggregated sparse tensor. Finally, after incorporating the prior information from group 1, the updated sparse tensor is used by the L-DFA module and sparse convolutional layers to predict the probability distribution of residuals for the voxels of group 2.\par
To effectively extract a large-scale context, a large receptive field is required. While regular convolutions can achieve this by stacking layers to expand the receptive field, sparse convolutions face a limitation due to their operation on only the occupied voxels. The presence of numerous empty voxels hinders the propagation of information across the field. Consequently, we have chosen to directly increase the kernel size of our convolution to 9, allowing for a broader gathering of information. In the architecture of our deep feature aggregation module, we employ dilated convolutions to construct an Inception-ResNet (IRN) module. This module is integrated into the backbone of the Deep Feature Aggregation (DFA) module~\cite{wang2022sparse}, further extending the receptive field. \par
In our proposed level-wise S2C context model, we have developed the Geometry Residual Coding (GRC) to prevent the decrease of the receptive field. Following that, we design the Residual Probability Approximation (RPA) model to further increase the receptive field, allowing for the extraction of a large-scale context. In addition, the grouped residual context model enables a two-stage parallel decoding approach, which is more efficient than the method involving up-sampling and eight-stage decoding as referenced in~\cite{wang2022sparse}.

\section{Experiments}
\label{sec: experiments}
In this section, we begin by introducing the experimental settings including the datasets, the evaluation metric, and implementation details. Subsequently, we show the point cloud compression performance of our proposed method. Furthermore, we conduct an in-depth ablation study of the design factors of our method to analyze their individual impacts.

\subsection{Experimental Settings}
\label{sec: experimental settings}
\subsubsection{\textbf{Datasets}}

The ShapeNet dataset~\cite{chang2015shapenet}, a comprehensive repository of 3D models, is used to train our proposed stage-wise S2C context model for dense point clouds. We employ the preprocessed and augmented dataset provided by~\cite{wang2022sparse} as the training dataset. For the testing phase, we evaluate our model on two publicly available datasets, i.e., MPEG 8i~\cite{d20178i} and Owlii~\cite{Xu2017Owlii}. The bit depth of the point cloud in MEPG 8i and Owlii is 10 and 11, respectively.
% follow the Common Test Condition (CTC)~\cite{iso2018mpeg} to 

We evaluate the performance of our proposed level-wise S2C context model on two sparse LiDAR datasets: the SemanticKITTI~\cite{behley2019semantickitti} dataset and the Ford~\cite{pandey2011ford} dataset. The KITTI dataset comprises 43,552 LiDAR scans derived from 22 point cloud sequences.
We follow the split in~\cite{wang2022sparse}, conducting training on sequences 00 to 11 and evaluation on sequences 12 to 21. The raw coordinates of the point cloud are quantized to 18 bits. The Ford dataset consists of 3 sequences, with each sequence containing 1500 frames. We follow the dataset partition in the MPEG standardization~\cite{iso2022mpeg}, training our model on sequence 01 and evaluating it on sequences 02 and 03.\par

\subsubsection{\textbf{Evaluation Metric}}
Given that the limitations of the receptive field become particularly pronounced at higher levels, we focus our evaluation of our S2C method on high-bit depth data. For dense point clouds with 10 or 11-bit depth,  we assess the lossless compression performance of our proposed stage-wise S2C context model. For sparse point clouds, the performance of point cloud compression methods is evaluated based on the rate-distortion trade-off. To quantify distortion, we utilize the point-to-point Peak Signal-to-Noise Ratio (D1 PSNR) and point-to-plane Peak Signal-to-Noise Ratio (D2 PSNR)~\cite{iso2022psnr} as metrics for geometry reconstruction quality. Specifically, we set the peak value of PSNR to 59.70 for the SemanticKITTI and 30000 for the Ford sequence.
We employ bit per point (bpp) as the rate metric, which is calculated by dividing the total length of the bitstream by the number of points.

\subsubsection{\textbf{Implementation Details}}
%量化
To employ our proposed level-wise S2C context model at more levels for sparse point clouds, we convert the point cloud coordinates from Cartesian to Spherical space. Subsequently, we adhere to the base quantization settings as detailed in~\cite{luo2024scp}, with the maximum bit depth set to 18. The range of the point cloud in Spherical coordinates is more compact compared to that in Cartesian coordinates, resulting in an actual maximum level of 16 for KITTI sequences and 17 for Ford sequences. We train our stage-wise S2C context model to perform coarse representation coding from level 1 to level 12. Our proposed Geometry Residual Coding method (GRC) method is applied starting from level 12, which has a point count approximately equal to the total point count. 
\par
%网络结构
In our proposed stage-wise S2C context model, we set the convolution kernel size to 3 and maintain the channel dimension C at 32 for dense point clouds. For sparse point clouds, the convolution kernel size is increased to 5. In our level-wise S2C context model, the Residual Probability Approximation (RPA) module acts as a large-scale context model. Within this module, the sparse convolution kernel size is set to 9, and the channel dimension C is configured to 32. The embedding layer is composed of a layer of convolution, and the Conv 1x1x1 layer comprises three sparse convolutions with a kernel size of 1.  \par
%训练，测试平台。
We train our proposed stage-wise S2C context model on the ShapeNet dataset with a batch size of 8 for 40 epochs, with the learning rate starting at $8\times10^{-4}$ and decreasing to $2\times10^{-5}$.
For the level-wise S2C context model on KITTI and Ford sequences with a batch size of 2 for 10 and 150 epochs, respectively. The learning rate for the Adam optimizer~\cite{kingma2014adam} is initially set at $8\times10^{-4}$ and is progressively reduced to $2.5\times10^{-5}$.   All the testing experiments are implemented on an NVIDIA A40 GPU. 

\begin{table*}[thb]
\caption{Bpp performance gains of lossless compression compared to G-PCC and SparsePCGC anchors on MPEG 8i and Owlii datasets.}
\label{tab:dense_main}
\centering
\scalebox{1.3}{
\begin{threeparttable}
\begin{tabular}{ccccccc}
\toprule
\multirow{2}{*}{\textbf{Dataset}} &
  \multirow{2}{*}{\textbf{Frame}} &
  \multirow{2}{*}{\textbf{G-PCC}} &
  \multirow{2}{*}{\textbf{\begin{tabular}[c]{@{}c@{}}SparsePCGC \end{tabular}}} &
  \multirow{2}{*}{\textbf{Ours}} &
  \multirow{2}{*}{\textbf{\begin{tabular}[c]{@{}c@{}}Gain over\\   G-PCC\end{tabular}}} &
  \multirow{2}{*}{\textbf{\begin{tabular}[c]{@{}c@{}}Gain over \\ SparsePCGC\end{tabular}}} \\
                         &                   &       &       &                 &      &      \\ \hline
\multirow{5}{*}{MPEG 8i} & longdress\_vox10  & 1.015 & 0.624 & \textbf{0.605} & -40.4\% & -3.0\% \\
                         & loot\_vox10       & 0.970 & 0.596 & \textbf{0.575} & -40.7\% & -3.5\% \\
                         & red\&black\_vox10 & 1.100 & 0.694 & \textbf{0.676} & -38.3\% & -2.6\% \\
                         & soldier\_vox10    & 1.030 & 0.630 & \textbf{0.612} & -40.6\% & -2.9\% \\
                         & queen\_vox10      & 0.773 & 0.511 & \textbf{0.478} & -38.2\% & -6.5\% \\ \hline
\multirow{2}{*}{Owlii}   & player\_vox11     & 0.898 & 0.506 & \textbf{0.483} & -46.2\% & -4.5\% \\
                         & dancer\_vox11     & 0.880 & 0.501 & \textbf{0.479} & -45.6\% & -4.3\% \\ \hline
\multicolumn{2}{l}{\textbf{Average}}         & 0.952 & 0.580 & \textbf{0.558} & -41.4\% & -3.8\% \\ \bottomrule
\end{tabular}
\end{threeparttable}}
\end{table*}

\par

\subsubsection{\textbf{Baselines}}
We use the voxel-based state-of-the-art algorithm, SparsePCGC~\cite{wang2022sparse} and the reference software TMC13-v14 of G-PCC~\cite{g-pccmpeg} as our baselines to verify the effectiveness of our S2C method. We have attempted to reproduce the SparsePCGC with a comparable or better performance because the official implementation is not available. Additionally, We train a Spherical-Coordinate-based SparsePCGC~\cite{wang2022sparse} to assess the impact of different coordinate systems on the performance of SparsePCGC. 

\subsection{Quantative Evaluation}
% sparse and dense point cloud lossless/near-lossless table
% R-D curve 
% sparse 每一层的码率分布，
% 编解码时间

\begin{table}[]
\caption{Lossless Encoding/decoding time (in seconds) on dense point cloud in MPEG 8i and Owlii.}
\label{tab:time_dense}
\centering
\setlength{\tabcolsep}{3.5mm}{
\scalebox{1.05}{
\begin{threeparttable}
\begin{tabular}{c|cc|cc}
\toprule
\multirow{2}{*}{method} & \multicolumn{2}{c|}{MPEG 8i(s)} & \multicolumn{2}{c}{Owlii(s)} \\
                        & Enc            & Dec            & Enc           & Dec          \\ \midrule
G-PCC                   & 4.59              & 3.02              & 16.69             & 9.73            \\ \hline
SparsePCGC              & 1.60              & 1.29              & 3.72             & 3.17            \\ \hline
Ours    & 1.16              & 1.06              & 2.84             & 2.68            \\ \bottomrule
\end{tabular}
\end{threeparttable}
}}
\end{table}

\textbf{Performance of stage-wise S2C context model.} 
In this experiment, we evaluate the lossless compression performance of our proposed stage-wise S2C context model, the reproduced SparsePCGC~\cite{wang2022sparse}, and G-PCC~\cite{g-pccmpeg} on dense object point clouds. The results are summarized in~\Cref{tab:dense_main}. Our method achieves state-of-the-art performance, outperforming SparsePCGC with a bit savings of -3.8\%. This indicates that our channel-wise autoregressive approach is more efficient in integrating and utilizing the label information from former stages compared to the point-wise autoregressive method~\cite{wang2022sparse}. Additionally, removing the upsampling operation allows for an expanded receptive field of sparse convolution on the encoding/decoding voxel, enabling the extraction of more informative context. We observe that our method achieves larger gains of compression performance with Owlii~\cite{Xu2017Owlii} at an 11-bit depth compared to MPEG 8i~\cite{d20178i} at a 10-bit depth. This improvement is attributed to the fact that the limitations of the receptive field become more pronounced at higher levels.
% Additionally, we evaluate the performance of our proposed method for lossy compression of low-level point clouds and observe that it results in only marginal or no improvements.
\par
We present the inference time of our stage-wise S2C context model on dense point clouds, in comparison with G-PCC~\cite{g-pccmpeg} and SparsePCGC~\cite{wang2022sparse}, as detailed in~\Cref{tab:time_dense}. It is observed that our stage-wise S2C context model achieves over 20\% and 15\% time savings for encoding and decoding, respectively. This efficiency is attributed to our channel-wise autoregressive approach, which preserves the number of processed points at each stage by storing the information of the previous stages in the channel dimension. In contrast, the point-wise autoregressive method~\cite{wang2022sparse} increases the number of points after each stage, leading to additional, unnecessary computational overhead.

% 0.54246+1.63134  0.510739+1.505852  0.35364+1.3233444   0.33665+1.219736 

\begin{figure*}[t]
\centering
\subfloat[PSNR D1 results on KITTI.]
{\includegraphics[width=3.4in]{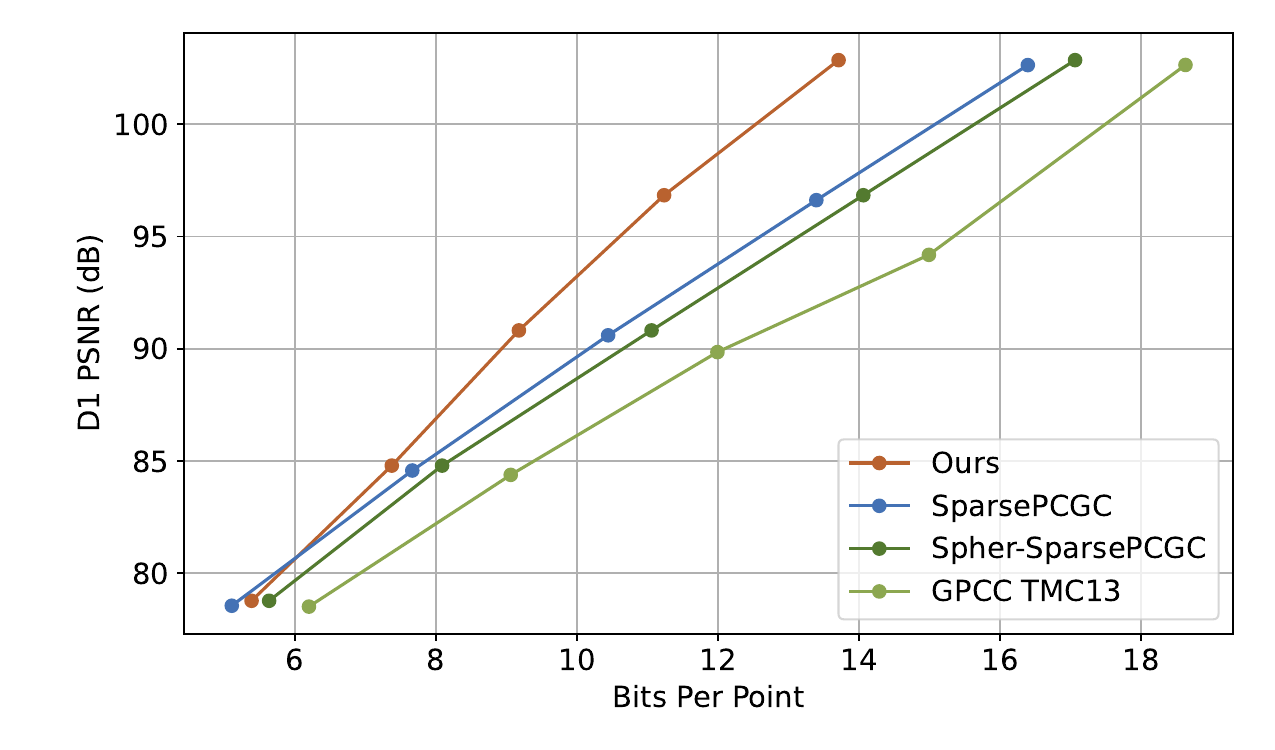}%
\label{kitti_1}}
\hfil
\subfloat[PSNR D1 results on Ford.]{\includegraphics[width=3.4in]{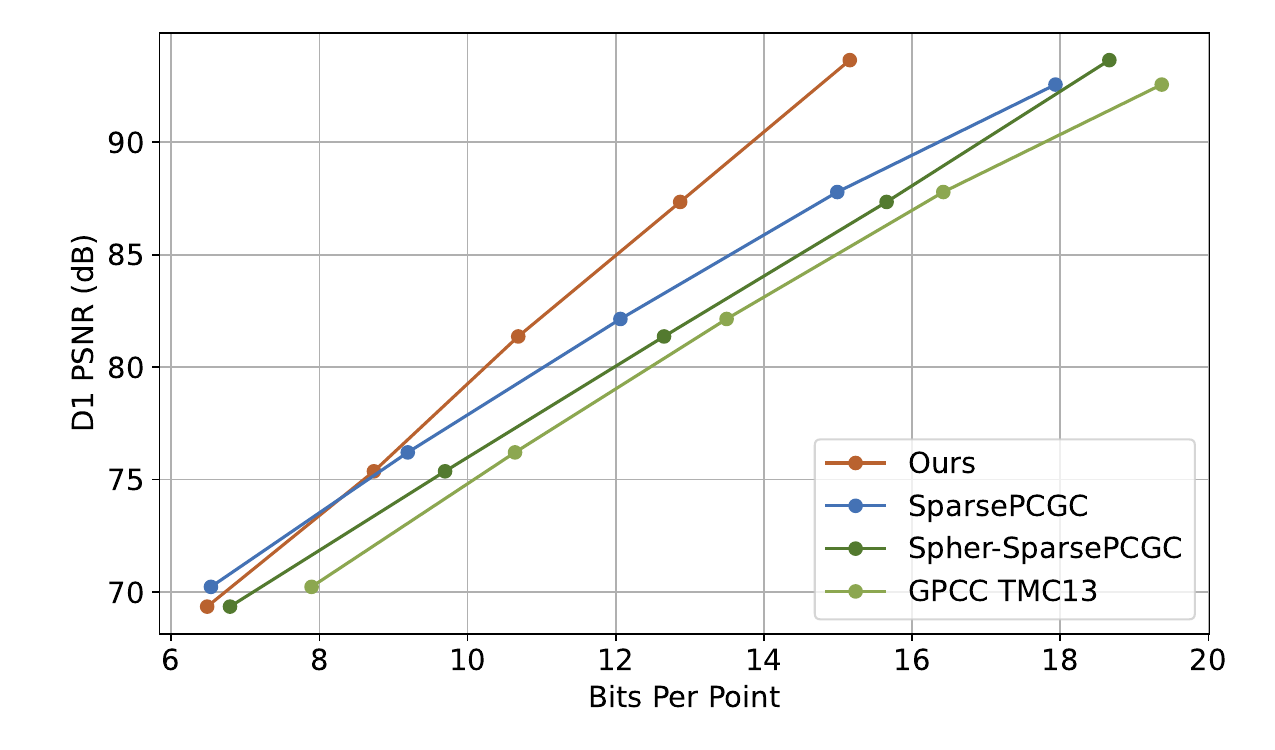}%
\label{ford_1}}
\hfil

\vspace{0.1em}
\subfloat[PSNR D2 results on KITTI.]{\includegraphics[width=3.4in]{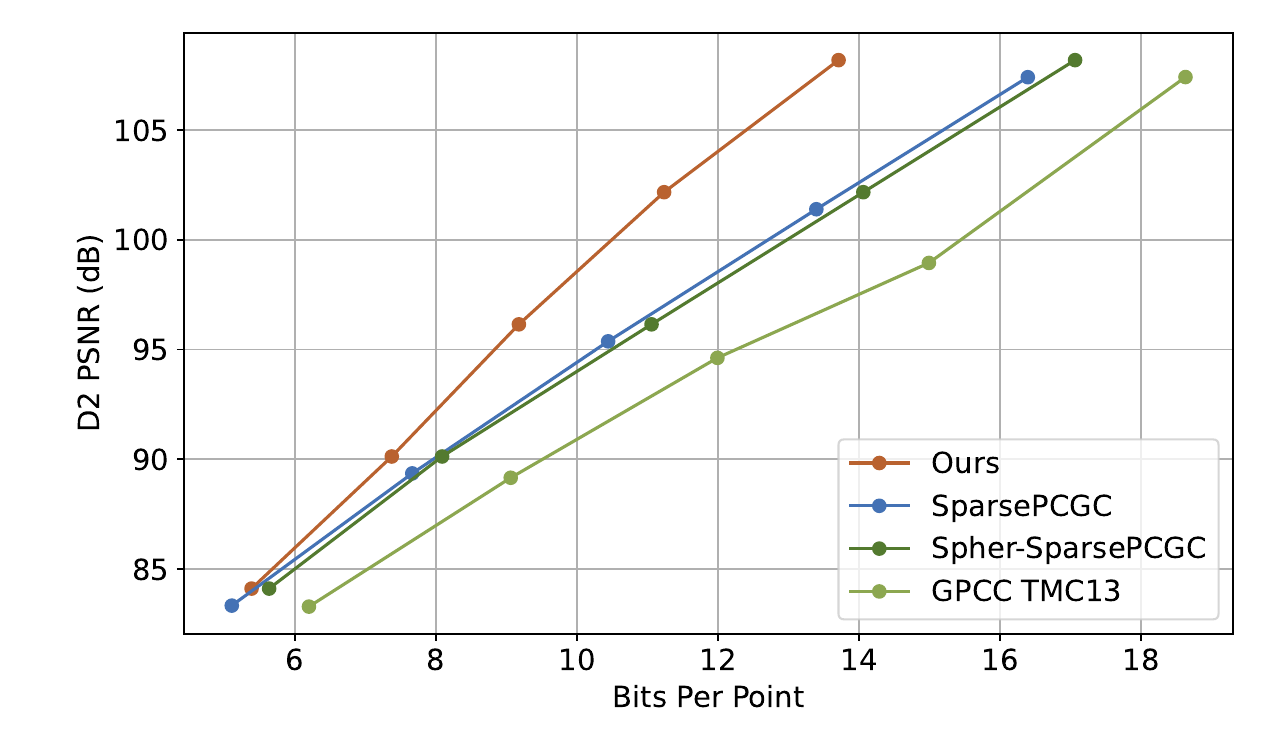}%
\label{kitti_2}}
\hfil
\subfloat[PSNR D2 results on Ford.]{\includegraphics[width=3.4in]{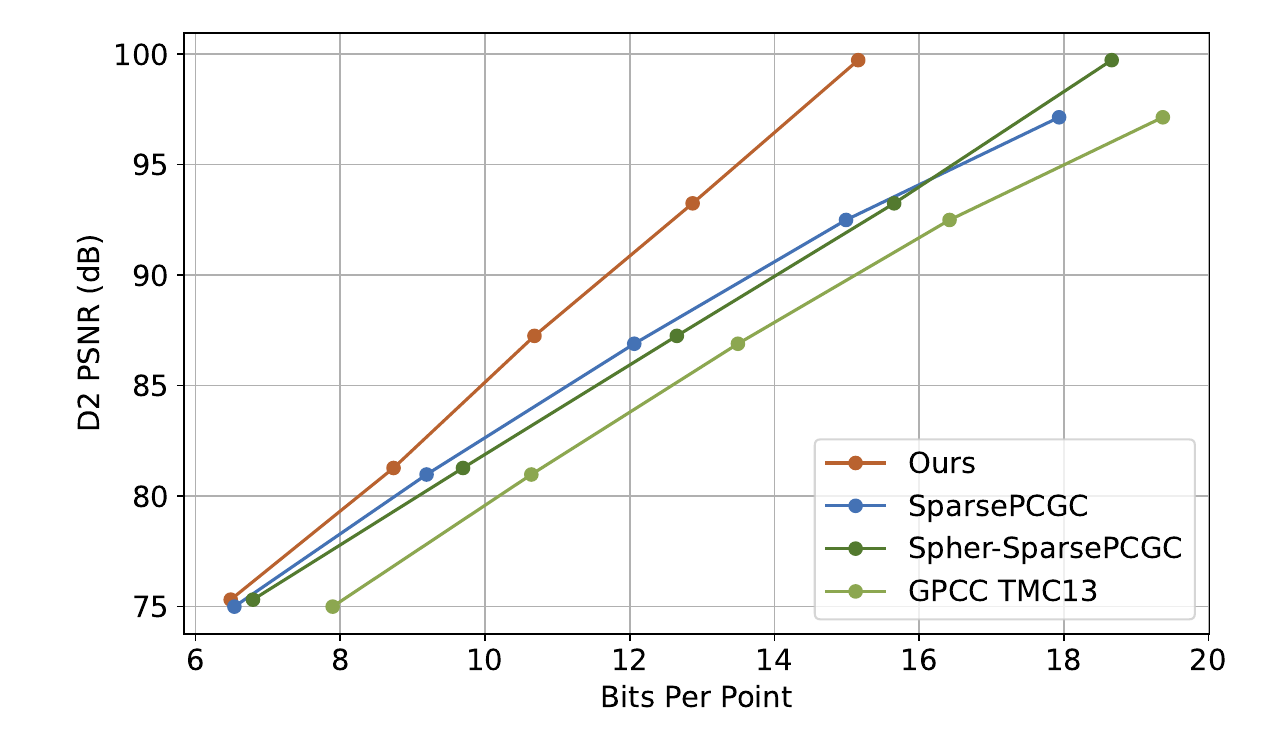}%
\label{ford_2}}
\hfil
\caption{Quantitative rate-distortion results of our proposed level-wise S2C context model on KITTI and Ford datasets. The baselines are voxel-based methods SparsePCGC~\cite{wang2022sparse}, Spher-SparsePCGC, and traditional codec G-PCC~\cite{g-pccmpeg}.}
\label{fig:sparse_main}
\end{figure*}

\begin{table}[]
\caption{BPP performance of lossless compression on 18-bit KITTI dataset in Cartesian coordinate system}
\label{tab:lossless}
\centering
\setlength{\tabcolsep}{1.15mm}{
\scalebox{1.05}{
\begin{threeparttable}

\begin{tabular}{c|c|ccc|cll}
\toprule
\multirow{2}{*}{Lossless} & \multirow{2}{*}{SparsePCGC} & \multicolumn{3}{c|}{Ours (level 15)}                    & \multicolumn{3}{c}{Ours (level 16)}                     \\
                        &                             & stage                 & level                & points & stage                  & level                & points \\ \midrule
\multirow{2}{*}{bpp}    & \multirow{2}{*}{16.39}     & \multicolumn{1}{c|}{7.47} & \multicolumn{1}{c|}{7.31} & 0.46  & \multicolumn{1}{c|}{10.18} & \multicolumn{1}{l|}{5.39} & 0.02  \\ \cline{3-8} 
                        &                             & \multicolumn{3}{c|}{15.24}                                      & \multicolumn{3}{c}{15.59}                                        \\ \bottomrule
\end{tabular}
\end{threeparttable}
}}
\end{table}

\textbf{Performance of level-wise S2C context model.}
We conduct experiments on sparse point cloud datasets to evaluate the compression performance of our proposed level-wise S2C context model. 
The Rate-Distortion (R-D) curve results are presented in Fig.~\ref{fig:sparse_main}. Our method is implemented in Spherical coordinate system, hence, we also test the Spher-SparsePCGC algorithm for comparison. It is observed that the compression performance of SparsePCGC in the Spherical coordinate system is inferior to that in the Cartesian coordinate system. This difference can be attributed to the complex distribution pattern of sparse point clouds in the Spherical coordinate system. However, voxel-based compression methods both struggle to capture and analyze the neighborhood patterns effectively due to their limited receptive fields and small parameter sizes. Our level-wise S2C context model starts from level 12 of Spher-SparsePCGC. We can observe that our method can outperform the efficient voxel-based method SparsePCGC with the level increasing. This is because our method can maintain the resolution at level 12 and design the efficient residual probability approximation module to extract a large-scale context. At the highest bit rate, our method can achieve 16.4\% and 15.4\% bit savings compared to SparsePCGC on KITTI and Ford datasets, respectively. \par   
It can not perform lossless compression in the Spherical coordinate system due to the precision error inherent in the space transformations. Consequently, we evaluate the lossless compression performance of our proposed method employed in Cartesian space and list the results in~\Cref{tab:lossless}. The bitstream comprises stage-wise coarse representation, level-wise residual, and directly encoded points. Level 15 and level 16 represent the start level of the level-wise context model. For our proposed S2C method at level 15, the coarse representation from level 1 to level 15 and the residual coding from level 16 to level 18 consume 7.47 bpp and 7.31 bpp, respectively.  Additionally, approximately a thousand points are encoded directly into the bitstream, requiring 0.46 bpp.
Our proposed method demonstrates the ability to achieve more than 1 bpp in bit savings compared to SparsePCGC~\cite{wang2022sparse} when performing lossless compression on the 18-bit KITTI dataset. Additionally, our level-wise context model demonstrates superior efficiency compared stage-wise context model when processing the sparser levels of LiDAR point clouds.
\par
% %比octree-based method 差，分析原因
% We also present the performance of the state-of-the-art Octree-based methods, SCP-EHEM~\cite{luo2024scp} and EHEM~\cite{song2023efficient}, in \Cref{fig:sparse_main}. Although our method achieves comparable performance with EHEM at high levels, it remains inferior to the SCP-EHEM. This disparity can be attributed to two main factors. Firstly, Octree-based methods process the Octree nodes directly through attention modules, which are not resolution-sensitive and well-suited for organizing large-scale sparse point clouds. Secondly, EHEM incorporates complex feature extraction modules and designs large-scale context extraction modules, enhancing compression performance but at the cost of significant parameter sizes and heavy computation. In contrast, this paper focuses on the resolution problem struggled by the voxel-based method and proposes a simple and efficient Geometry Residual Coding (GRC) technique to significantly expand the receptive field. 
\par

\begin{table}[]
\caption{Lossy Encoding/decoding time (in seconds) on sparse point cloud. The quantized bit depth is 17 for KITTI and Ford.}
\label{tab:time_sparse}
\centering
\setlength{\tabcolsep}{3.5mm}{
\scalebox{1.05}{
\begin{threeparttable}
\begin{tabular}{c|cc|cc}
\toprule
\multirow{2}{*}{method} & \multicolumn{2}{c|}{KITTI(s)} & \multicolumn{2}{c}{Ford(s)} \\
                        & Enc           & Dec           & Enc          & Dec          \\ \midrule
G-PCC                   & 0.85             & 0.57             & 1.00            & 0.63            \\ \hline
SparsePCGC              & 2.38             & 1.99             & 2.03            & 1.61            \\ \hline
Spher-SparsePCGC        & 2.43             & 1.99             & 1.92            & 1.50            \\ \hline
Ours    & 2.17       & 2.02           & 1.68            & 1.56           \\ \bottomrule
\end{tabular}
\end{threeparttable}
}}
\end{table}

We present the inference time of our proposed S2C context model on sparse point clouds, as detailed~\Cref{tab:time_sparse}. As shown in~\Cref{tab:time_sparse}, our S2C context model achieves an 8.8\%$\sim$17.2\% reduction in encoding time and maintains comparable decoding time compared to the voxel-based SparsePCGC. This improvement stems from the elimination of the serial eight-stage autoregressive context in~\cite{wang2022sparse} and the adoption of a two-stage grouped residual context at high levels in our level-wise S2C context model, which substantially decreases the model's inference time. 

\begin{figure*}[t]
\centering
\centerline{\includegraphics[width=\linewidth]{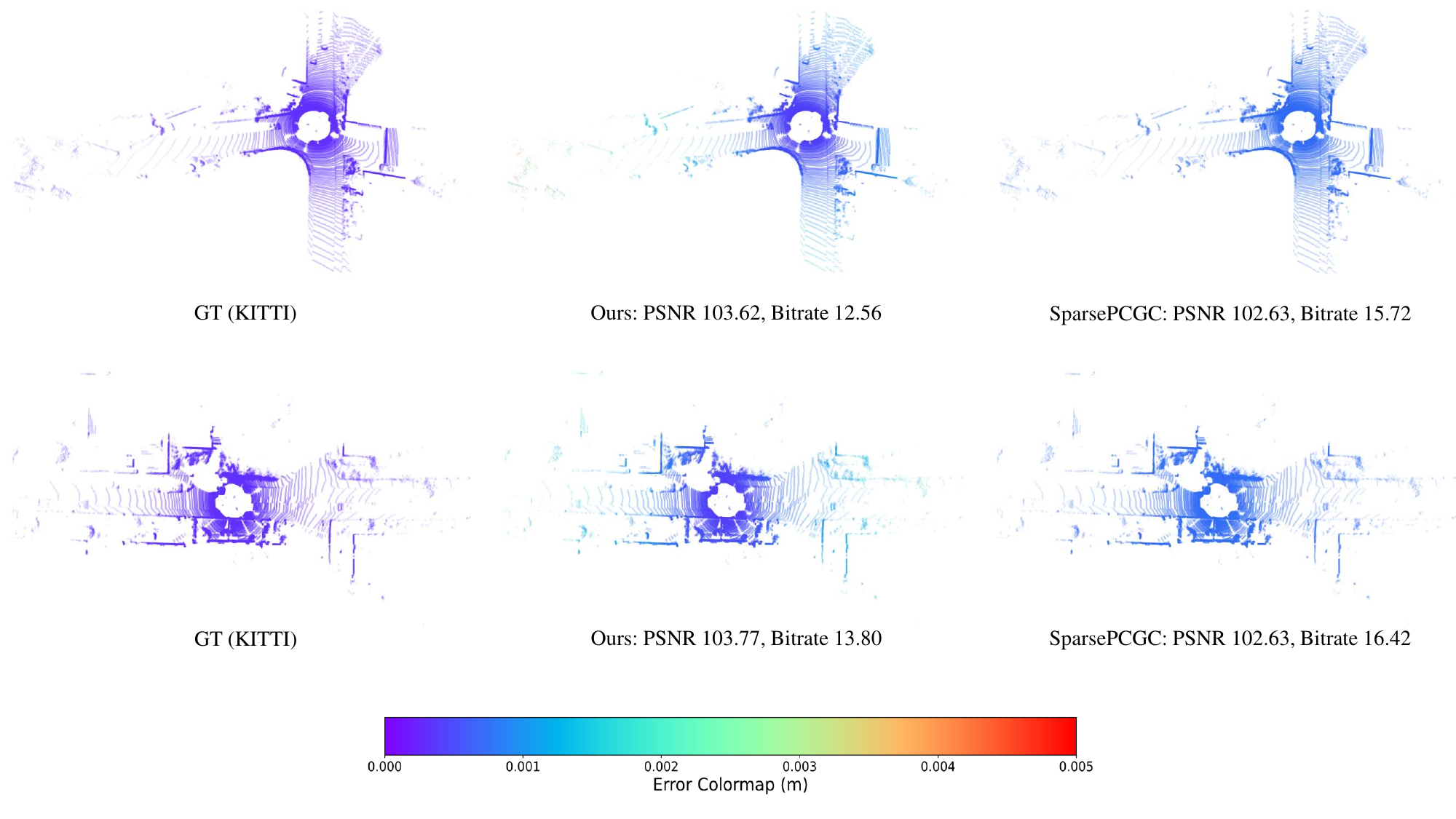}}
\caption{Qualitative reconstruction comparisons of No.000000 in Sequence 11 (top) and No.000000 in Sequence 15 (bottom) from KITTI dataset~\cite{behley2019semantickitti} for ground truth, ours, and SparsePCGC~\cite{wang2022sparse}. The error colormap denotes the point-to-point distances in meters (m). D1 PSNR and bitrate metrics are also depicted.
}
\label{vis}
\end{figure*}

\subsection{Qualitative Analysis}
%kitti的可视化
Fig.~\ref{vis} presents a qualitative comparison of our proposed method against the results of the baseline voxel-based algorithm SparsePCGC~\cite{wang2022sparse} on the KITTI dataset~\cite{behley2019semantickitti}. It is evident that the reconstruction quality of our method surpasses that of SparsePCGC. This enhancement can be attributed to the use of  Spherical coordinates, which allow for denser division of sub-spaces in the vicinity and sparser division in distant regions, aligning with the distribution of point clouds collected by rotating LiDAR systems. Consistent with this division, we observe that the error in distance increases proportionally with the distance value in Spherical coordinates. This observation is corroborated by the findings in ~\cite{luo2024scp}, which employs multi-level point cloud partitioning to mitigate this issue. Furthermore, our method achieves a reduced bitrate through the application of geometry residual coding and a residual probability approximation module, enabling the extraction of large-scale informative context. 

\begin{table}[]
\caption{Comparison of bpp results between different kernel sizes of sparse convolution on Ford. The quantized bit depth is 17.}
\centering
\label{tab:ab_kernel}
\setlength{\tabcolsep}{4mm}{
\scalebox{1.05}{
\begin{threeparttable}
\begin{tabular}{c|cccc}
\toprule
Kernel size & 5 & 7 & 9 & 11 \\ \midrule
Bpp         & 16.33 & 15.65 & 15.19 & 14.62  \\ \bottomrule
\end{tabular}
\end{threeparttable}
}}
\end{table}

\begin{table}[]
\caption{Comparison of bpp results between different channel numbers of sparse convolution on MPEG 8i.}
\centering
\label{tab:ab_channel}
\setlength{\tabcolsep}{6.2mm}{
\scalebox{1.05}{
\begin{threeparttable}
\begin{tabular}{c|ccc}
\toprule
Channel & 16 & 32 & 64 \\ \midrule
Bpp     & 0.606  & 0.589  & 0.587  \\ \bottomrule
\end{tabular}
\end{threeparttable}
}}
\end{table}

\subsection{Ablation Study}
%空间坐标系/层数  %残差编码  %大核卷积&空洞卷积5，7，9 
To assess the impact of the receptive field on the performance of voxel-based methods for sparse point clouds, we perform ablation experiments on the kernel size of sparse convolution within our proposed level-wise context model, with results detailed in~\Cref{tab:ab_kernel}. We observe that the bits per point (bpp) decrease as the kernel size increases, indicating that a larger kernel size effectively captures neighborhood information, thereby accurately predicting the residual position. This observation underscores the voxel-based method's ongoing challenge with the limitation of the receptive field when processing large-scale sparse point clouds, suggesting potential avenues for further improvement. \par
For dense point clouds, we conduct ablation experiments on the number of channels in sparse convolution within our proposed stage-wise context model, with results presented in~\Cref{tab:ab_channel}. We find that an increased number of channels enhances the model's information integration ability, leading to a reduction in bpp. However, a higher number of channels also correlates with an increased inference time. Consequently, we set the base channel number to 32.

\section{Conclusion}
\label{sec: conclusion}
In this paper, we introduce an efficient Space-to-Channel (S2C) approach to enhance the analytical capabilities of sparse convolution for point cloud geometry compression. Building on this concept, We propose stage-wise and level-wise S2C context models that offer a larger receptive field and reduced computational complexity for dense and sparse point clouds, respectively. Experimental results demonstrate that our S2C context model achieves superior compression performance and reduced codec time across both two types of point clouds compared to the state-of-the-art voxel-based methods. Our work advances the utilization of sparse convolution and provides a valuable reference for further research and development of voxel-based point cloud compression in a unified framework.

\bibliographystyle{IEEEtran}
\bibliography{paper}

\begin{thebibliography}{10}
\providecommand{\url}[1]{#1}
\csname url@samestyle\endcsname
\providecommand{\newblock}{\relax}
\providecommand{\bibinfo}[2]{#2}
\providecommand{\BIBentrySTDinterwordspacing}{\spaceskip=0pt\relax}
\providecommand{\BIBentryALTinterwordstretchfactor}{4}
\providecommand{\BIBentryALTinterwordspacing}{\spaceskip=\fontdimen2\font plus
\BIBentryALTinterwordstretchfactor\fontdimen3\font minus \fontdimen4\font\relax}
\providecommand{\BIBforeignlanguage}[2]{{%
\expandafter\ifx\csname l@#1\endcsname\relax
\typeout{** WARNING: IEEEtran.bst: No hyphenation pattern has been}%
\typeout{** loaded for the language `#1'. Using the pattern for}%
\typeout{** the default language instead.}%
\else
\language=\csname l@#1\endcsname
\fi
#2}}
\providecommand{\BIBdecl}{\relax}
\BIBdecl

\bibitem{krivokuca2019volumetric}
M.~Krivoku{\'c}a, P.~A. Chou, and M.~Koroteev, ``A volumetric approach to point cloud compression--part ii: Geometry compression,'' \emph{IEEE Transactions on Image Processing}, vol.~29, pp. 2217--2229, 2019.

\bibitem{tzamarias2022fast}
D.~E. Tzamarias, K.~Chow, I.~Blanes, and J.~Serra-Sagrist{\`a}, ``Fast run-length compression of point cloud geometry,'' \emph{IEEE transactions on image processing}, vol.~31, pp. 4490--4501, 2022.

\bibitem{thanou2016graph}
D.~Thanou, P.~A. Chou, and P.~Frossard, ``Graph-based compression of dynamic 3d point cloud sequences,'' \emph{IEEE Transactions on Image Processing}, vol.~25, no.~4, pp. 1765--1778, 2016.

\bibitem{de2016compression}
R.~L. De~Queiroz and P.~A. Chou, ``Compression of 3d point clouds using a region-adaptive hierarchical transform,'' \emph{IEEE Transactions on Image Processing}, vol.~25, no.~8, pp. 3947--3956, 2016.

\bibitem{g-pccmpeg}
``Mpeg g-pcc tmc13. https://github.com/mpeggroup/mpeg-pcc-tmc13.'' \emph{ISO/IEC JTC 1/SC 29/WC 7}, 2021.

\bibitem{jackins1980oct}
C.~L. Jackins and S.~L. Tanimoto, ``Oct-trees and their use in representing three-dimensional objects,'' \emph{Computer Graphics and Image Processing}, vol.~14, no.~3, pp. 249--270, 1980.

\bibitem{meagher1982geometric}
D.~Meagher, ``Geometric modeling using octree encoding,'' \emph{Computer graphics and image processing}, vol.~19, no.~2, pp. 129--147, 1982.

\bibitem{schnabel2006octree}
R.~Schnabel and R.~Klein, ``Octree-based point-cloud compression.'' \emph{PBG@ SIGGRAPH}, vol.~3, pp. 111--121, 2006.

\bibitem{luo2024scp}
A.~Luo, L.~Song, K.~Nonaka, K.~Unno, H.~Sun, M.~Goto, and J.~Katto, ``Scp: Spherical-coordinate-based learned point cloud compression,'' in \emph{Proceedings of the AAAI Conference on Artificial Intelligence}, vol.~38, no.~4, 2024, pp. 3954--3962.

\bibitem{wang2022sparse}
J.~Wang, D.~Ding, Z.~Li, X.~Feng, C.~Cao, and Z.~Ma, ``Sparse tensor-based multiscale representation for point cloud geometry compression,'' \emph{IEEE Transactions on Pattern Analysis and Machine Intelligence}, vol.~45, no.~7, pp. 9055--9071, 2022.

\bibitem{fu2022octattention}
C.~Fu, G.~Li, R.~Song, W.~Gao, and S.~Liu, ``Octattention: Octree-based large-scale contexts model for point cloud compression,'' in \emph{Proceedings of the AAAI conference on artificial intelligence}, vol.~36, no.~1, 2022, pp. 625--633.

\bibitem{song2023efficient}
R.~Song, C.~Fu, S.~Liu, and G.~Li, ``Efficient hierarchical entropy model for learned point cloud compression,'' in \emph{Proceedings of the IEEE/CVF Conference on Computer Vision and Pattern Recognition}, 2023, pp. 14\,368--14\,377.

\bibitem{vaswani2017attention}
A.~Vaswani, ``Attention is all you need,'' \emph{Advances in Neural Information Processing Systems}, 2017.

\bibitem{wang2021multiscale}
J.~Wang, D.~Ding, Z.~Li, and Z.~Ma, ``Multiscale point cloud geometry compression,'' in \emph{2021 Data Compression Conference (DCC)}.\hskip 1em plus 0.5em minus 0.4em\relax IEEE, 2021, pp. 73--82.

\bibitem{choy20194d}
C.~Choy, J.~Gwak, and S.~Savarese, ``4d spatio-temporal convnets: Minkowski convolutional neural networks,'' in \emph{Proceedings of the IEEE Conference on Computer Vision and Pattern Recognition}, 2019, pp. 3075--3084.

\bibitem{schwarz2018emerging}
S.~Schwarz, M.~Preda, V.~Baroncini, M.~Budagavi, P.~Cesar, P.~A. Chou, R.~A. Cohen, M.~Krivoku{\'c}a, S.~Lasserre, Z.~Li \emph{et~al.}, ``Emerging mpeg standards for point cloud compression,'' \emph{IEEE Journal on Emerging and Selected Topics in Circuits and Systems}, vol.~9, no.~1, pp. 133--148, 2018.

\bibitem{garcia2019geometry}
D.~C. Garcia, T.~A. Fonseca, R.~U. Ferreira, and R.~L. De~Queiroz, ``Geometry coding for dynamic voxelized point clouds using octrees and multiple contexts,'' \emph{IEEE Transactions on Image Processing}, vol.~29, pp. 313--322, 2019.

\bibitem{cao2021compression}
C.~Cao, M.~Preda, V.~Zakharchenko, E.~S. Jang, and T.~Zaharia, ``Compression of sparse and dense dynamic point clouds—methods and standards,'' \emph{Proceedings of the IEEE}, vol. 109, no.~9, pp. 1537--1558, 2021.

\bibitem{huang2020octsqueeze}
L.~Huang, S.~Wang, K.~Wong, J.~Liu, and R.~Urtasun, ``Octsqueeze: Octree-structured entropy model for lidar compression,'' in \emph{Proceedings of the IEEE/CVF conference on computer vision and pattern recognition}, 2020, pp. 1313--1323.

\bibitem{que2021voxelcontext}
Z.~Que, G.~Lu, and D.~Xu, ``Voxelcontext-net: An octree based framework for point cloud compression,'' in \emph{Proceedings of the IEEE/CVF Conference on Computer Vision and Pattern Recognition}, 2021, pp. 6042--6051.

\bibitem{quach2019learning}
M.~Quach, G.~Valenzise, and F.~Dufaux, ``Learning convolutional transforms for lossy point cloud geometry compression,'' in \emph{2019 IEEE international conference on image processing (ICIP)}.\hskip 1em plus 0.5em minus 0.4em\relax IEEE, 2019, pp. 4320--4324.

\bibitem{quach2020improved}
------, ``Improved deep point cloud geometry compression,'' in \emph{2020 IEEE 22nd International Workshop on Multimedia Signal Processing (MMSP)}.\hskip 1em plus 0.5em minus 0.4em\relax IEEE, 2020, pp. 1--6.

\bibitem{wang2021lossy}
J.~Wang, H.~Zhu, H.~Liu, and Z.~Ma, ``Lossy point cloud geometry compression via end-to-end learning,'' \emph{IEEE Transactions on Circuits and Systems for Video Technology}, vol.~31, no.~12, pp. 4909--4923, 2021.

\bibitem{gu20193d}
S.~Gu, J.~Hou, H.~Zeng, H.~Yuan, and K.-K. Ma, ``3d point cloud attribute compression using geometry-guided sparse representation,'' \emph{IEEE Transactions on Image Processing}, vol.~29, pp. 796--808, 2019.

\bibitem{pang2022grasp}
J.~Pang, M.~A. Lodhi, and D.~Tian, ``Grasp-net: Geometric residual analysis and synthesis for point cloud compression,'' in \emph{Proceedings of the 1st International Workshop on Advances in Point Cloud Compression, Processing and Analysis}, 2022, pp. 11--19.

\bibitem{pandey2011ford}
G.~Pandey, J.~R. McBride, and R.~M. Eustice, ``Ford campus vision and lidar data set,'' \emph{The International Journal of Robotics Research}, vol.~30, no.~13, pp. 1543--1552, 2011.

\bibitem{thomas2018semantic}
H.~Thomas, F.~Goulette, J.-E. Deschaud, B.~Marcotegui, and Y.~LeGall, ``Semantic classification of 3d point clouds with multiscale spherical neighborhoods,'' in \emph{2018 International conference on 3D vision (3DV)}.\hskip 1em plus 0.5em minus 0.4em\relax IEEE, 2018, pp. 390--398.

\bibitem{lei2020spherical}
H.~Lei, N.~Akhtar, and A.~Mian, ``Spherical kernel for efficient graph convolution on 3d point clouds,'' \emph{IEEE transactions on pattern analysis and machine intelligence}, vol.~43, no.~10, pp. 3664--3680, 2020.

\bibitem{zhang2020polarnet}
Y.~Zhang, Z.~Zhou, P.~David, X.~Yue, Z.~Xi, B.~Gong, and H.~Foroosh, ``Polarnet: An improved grid representation for online lidar point clouds semantic segmentation,'' in \emph{Proceedings of the IEEE/CVF conference on computer vision and pattern recognition}, 2020, pp. 9601--9610.

\bibitem{behley2019semantickitti}
J.~Behley, M.~Garbade, A.~Milioto, J.~Quenzel, S.~Behnke, C.~Stachniss, and J.~Gall, ``Semantickitti: A dataset for semantic scene understanding of lidar sequences,'' in \emph{Proceedings of the IEEE/CVF international conference on computer vision}, 2019, pp. 9297--9307.

\bibitem{chang2015shapenet}
A.~X. Chang, T.~Funkhouser, L.~Guibas, P.~Hanrahan, Q.~Huang, Z.~Li, S.~Savarese, M.~Savva, S.~Song, H.~Su \emph{et~al.}, ``Shapenet: An information-rich 3d model repository,'' \emph{arXiv preprint arXiv:1512.03012}, 2015.

\bibitem{d20178i}
E.~d’Eon, B.~Harrison, T.~Myers, and P.~A. Chou, ``8i voxelized full bodies—a voxelized point cloud dataset,'' \emph{IEC JTC1/SC29 Joint WG11/WG1 (MPEG/JPEG) input document WG11M40059/WG1M74006, Geneva}, 2017.

\bibitem{Xu2017Owlii}
{Yi Xu, Yao Lu, Ziyu Wen}, ``{Owlii Dynamic human mesh sequence dataset},'' \emph{{ISO/IEC JTC1/SC29/WG11 m41658,120th MPEG Meeting, Macau}}, {October} {2017}.

\bibitem{iso2022mpeg}
``Mpeg 3d graphics coding. preliminary dataset for ai-based point cloud experiments.'' 2022.

\bibitem{iso2022psnr}
``Mpeg 3d graphics coding. common test conditions for g-pcc.'' 2021.

\bibitem{kingma2014adam}
D.~P. Kingma, ``Adam: A method for stochastic optimization,'' \emph{arXiv preprint arXiv:1412.6980}, 2014.

\end{thebibliography}

\end{document}